# Communications Inspired Linear Discriminant Analysis

**Minhua Chen** MINHUA.CHEN@DUKE.EDU
**William Carson**[†] WILLIAM.CARSON@PACONSULTING.COM
**Miguel Rodrigues**[‡] M.RODRIGUES@EE.UCL.AC.UK
**Robert Calderbank** ROBERT.CALDERBANK@DUKE.EDU
**Lawrence Carin** LCARIN@DUKE.EDU
Department of ECE, Duke University, Durham, NC, USA
[†]PA Consulting Group, Cambridge Technology Centre, Melbourn, UK
[‡]Department of E&EE, University College London, London, UK

## Abstract

We study the problem of supervised linear dimensionality reduction, taking an information-theoretic viewpoint. The linear projection matrix is designed by maximizing the mutual information between the projected signal and the class label. By harnessing a recent theoretical result on the gradient of mutual information, the above optimization problem can be solved directly using gradient descent, without requiring simplification of the objective function. Theoretical analysis and empirical comparison are made between the proposed method and two closely related methods, and comparisons are also made with a method in which Rényi entropy is used to define the mutual information (in this case the gradient may be computed simply, under a special parameter setting). Relative to these alternative approaches, the proposed method achieves promising results on real datasets.

## 1. Introduction

The analysis of high-dimensional data is of interest in many applications. To reduce the cost of data processing, and to increase the interpretability of the data, one typically employs dimensionality reduction as a pre-processing step. It also plays the role of regularization for the data. Although nonlinear dimensionality reduction methods (Tenenbaum et al., 2000; Song et al., 2007) have become popular recently, linear dimensionality reduction methods still play an important role, mainly due to their simplicity. Linear dimensionality reduction based on random projections has gained significant attention recently, as a result of success in compressive sensing (Candes & Wakin, 2008) and other applications (Liu & Fieguth, 2012). However, random projections may not be the best choice if we know the statistical properties of the underlying signal (Duarte-Carvajalino & Sapiro, 2009). Hence, an important question to be answered is how to design the projection matrix so that the measurement is the most informative.

In this paper we focus on projection design for classification, or supervised dimensionality reduction. Linear Discriminant Analysis (LDA) (Fisher, 1936) is one of the most important supervised dimensionality reduction methods. The design criterion of LDA maximizes the between-class scattering while minimizing the within-class scattering of the projected data, with these two criteria addressed simultaneously. It has been proven that under mild conditions this criterion is Bayes optimal (Hamsici & Martinez, 2008). However, this method has two disadvantages. First, the dimensionality of the projected space in LDA can only be less than the number of data classes, which greatly restricts its applicability. Second, LDA only uses first and second order statistics of the data, ignoring higher-order information. To overcome these two disadvantages, other criteria have been proposed in the literature (Tao et al., 2009), out of which an important category is the information-theoretic criterion.

In the information-theoretic approach, the projection matrix is designed by maximizing the mutual information (MI) between the projected signal and the class label (Torkkola, 2003; 2001; Nenadic, 2007; Kaski & Peltonen, 2003; Hild et al., 2006). Intuitively, the larger the mutual information is, the better it is for the projected signal to recover the label information. Theoretically, the Bayes classification error is bounded by the MI (Nenadic, 2007) (based on a Shannon entropy measure). However, the MI is not easy to calculate, posing a significant obstacle to its optimization. Almost all existing information-theoretic-based algorithms seek an approximation to the Shannon MI, hence compromising the objective function. For example, in recent studies (Torkkola, 2003; 2001; Hild et al., 2006), the quadratic mutual information (with quadratic Rényi entropy) is used instead of the Shannon-based MI; this is because with the use of quadratic Rényi entropy, the gradient of MI can be calculated analytically under the assumption of a Gaussian mixture model (GMM) signal model. In the work of (Kaski & Peltonen, 2003), the Shannon MI is approximated by its empirical estimation on the training data. Using Information Discriminant Analysis (IDA) (Nenadic, 2007), the entropy of the GMM in the MI calculation, where the higher-order information comes into play, is approximated with the entropy of a global Gaussian distribution, which again loses the higher-order information. The LDA method, although not proposed under the information-theoretic criterion, can also be viewed as an approximation to the MI objective function.

The main contribution of this paper is to show that the use of Shannon MI optimization, for linear feature design in classification, can be solved directly, without compromising or simplifying the objective function. The key tool is a theoretical result that recently appeared in the communications literature, which gives an explicit expression for the gradient of Shannon MI with respect to the projection matrix in linear vector Gaussian channels (Palomar & Verdu, 2006). This theorem has found applications in the area of precoder design for communication systems (Xiao et al., 2011; Carson et al., 2012), but is not widely appreciated in the machine learning and signal processing communities, except for a few papers on optical imaging system design (Ashok et al., 2008; Baheti & Neifeld, 2009). Our paper is the first to apply this theorem to the supervised dimensionality reduction problem. As a result, we obtain a new explicit expression for the gradient of the Shannon MI objective function for *any* input signal distribution, which is not achieved in any of the methods mentioned above. Consequently, numerical optimization methods (*e.g.*, gradient descent) can be applied. Since we make no assumptions on the input signal distribution in each class, our analytical result is very general and can be applied to a broad spectrum of applications. Additionally, we perform a theoretical analysis of this design metric, providing new insights. To connect to the quadratic mutual information approach (Torkkola, 2003; 2001), we adopt the mixture-of-GMMs signal model.

## 2. Main Result

Suppose the label and data are generated i.i.d. via the following process: $c \sim \text{Mult}(c; 1, \boldsymbol{w})$; $\boldsymbol{x}|c \sim p(\boldsymbol{x}|c)$ where $\boldsymbol{w} \in \mathbb{R}^{M \times 1}$ is the prior distribution on the $M$ classes, $\boldsymbol{x} \in \mathbb{R}^{p \times 1}$ is the original signal, and $p(\boldsymbol{x}|c)$ is the data distribution for class $c$. Hence the joint density is $p(\boldsymbol{x}, c) = w_c p(\boldsymbol{x}|c)$, and the global signal density can be written as

$$p(\boldsymbol{x}) = \sum_{m=1}^{M} w_m p(\boldsymbol{x}|m). \tag{1}$$

Here we make no assumption on the form of $p(\boldsymbol{x}|m)$; hence the above signal model is very general. We do assume that $\boldsymbol{w}$ and $p(\boldsymbol{x}|m)$ are known or can be estimated from training data.

In supervised dimensionality reduction, we seek a projection matrix $\boldsymbol{\Phi} \in \mathbb{R}^{d \times p}$ such that the projected signal

$$\boldsymbol{y} = \boldsymbol{\Phi}\boldsymbol{x} + \boldsymbol{\epsilon} \tag{2}$$

is the most informative in identifying the underlying class label $c$. We assume the measurement noise $\boldsymbol{\epsilon}$ is Gaussian, i.e., $p(\boldsymbol{y}|\boldsymbol{x}) = \mathcal{N}(\boldsymbol{y}; \boldsymbol{\Phi}\boldsymbol{x}, \boldsymbol{R}^{-1})$ where $\boldsymbol{R}$ is the known noise precision matrix. We adopt the information-theoretic criterion (Nenadic, 2007) as

$$\max_{\boldsymbol{\Phi}} \quad I(C; \boldsymbol{Y}) \quad \text{s.t.} \quad \boldsymbol{\Phi}\boldsymbol{\Phi}^\top = \boldsymbol{I}_d \tag{3}$$

where $C$ and $\boldsymbol{Y}$ represent $c$ and $\boldsymbol{y}$ as random variables, $I(C; \boldsymbol{Y})$ denotes the MI, and the orthonormality constraint is common in the literature (Nenadic, 2007). Intuitively, the larger the MI is, the better it is for the projected signal $\boldsymbol{y}$ to predict the latent class label $c$. Theoretically, there is also a strong justification for the above criterion. The Bayes classification error, defined as $P_e = \int p(\boldsymbol{y})(1 - \max_c p(c|\boldsymbol{y}))d\boldsymbol{y}$, can be bounded by $I(C; \boldsymbol{Y})$ as follows (Hellman & Raviv, 1970; Fano, 1961; Nenadic, 2007)

$$\frac{H(C|\boldsymbol{Y}) - H(P_e)}{\log M} \leq P_e \leq \frac{1}{2} H(C|\boldsymbol{Y}) \tag{4}$$

where $H(C|\boldsymbol{Y}) = H(C) - I(C; \boldsymbol{Y})$ and $0 \leq H(P_e) \leq 1$. Hence, the smaller $H(C|\boldsymbol{Y})$ the tighter the bound will be for $P_e$, and minimizing $H(C|\boldsymbol{Y})$ corresponds to maximizing $I(C; \boldsymbol{Y})$. Note that (4) is based on a Shannon definition of entropy (*e.g.*, *not* a Rényi entropy measure (Torkkola, 2003), with a comparison to results based on Rényi entropy discussed below). Unless

stated otherwise, all measures of entropy and differential entropy discussed below are based on a Shannon definition (Cover & Thomas, 2006).

In order to solve the optimization problem in (3), we first introduce a theoretical result that appeared in the communications literature:

**Theorem 1.** *(Palomar & Verdu, 2006) Given the measurement model in (2), the gradient of mutual information $I(\boldsymbol{X};\boldsymbol{Y})$ with respect to the projection matrix $\boldsymbol{\Phi}$ can be expressed as*

$$\nabla_{\boldsymbol{\Phi}} I(\boldsymbol{X};\boldsymbol{Y}) = \boldsymbol{R}\boldsymbol{\Phi}\boldsymbol{\Sigma} \tag{5}$$

*where $\boldsymbol{\Sigma} = \int p(\boldsymbol{y}) \int p(\boldsymbol{x}|\boldsymbol{y})(\boldsymbol{x}-\boldsymbol{x_y})(\boldsymbol{x}-\boldsymbol{x_y})^\top d\boldsymbol{x}d\boldsymbol{y}$ is the MMSE matrix, and $\boldsymbol{x_y} = \int \boldsymbol{x}p(\boldsymbol{x}|\boldsymbol{y})d\boldsymbol{x}$ is the posterior mean.*

This theorem provides a connection between information theory and estimation theory, by linking the gradient of mutual information to the MMSE matrix. It has found applications in precoder design for communications systems (Xiao et al., 2011; Carson et al., 2012). However, the power of this theorem has not been widely applied in the machine learning and signal processing communities. The only studies we found are (Ashok et al., 2008; Baheti & Neifeld, 2009) which use the above theorem to design optical imaging systems. This paper is the first work to apply and extend the above theorem to the supervised linear-dimensionality-reduction problem. Our main result is summarized in the following new theorem:

**Theorem 2.** *Given the measurement model in (2) and the multi-class signal model in (1), the gradient of mutual information $I(C;\boldsymbol{Y})$ with respect to the projection matrix $\boldsymbol{\Phi}$ can be expressed as*

$$\nabla_{\boldsymbol{\Phi}} I(C;\boldsymbol{Y}) = \boldsymbol{R}\boldsymbol{\Phi}\widetilde{\boldsymbol{\Sigma}} \tag{6}$$

*with the equivalent MMSE matrix $\widetilde{\boldsymbol{\Sigma}}$ expressed as*

$$\widetilde{\boldsymbol{\Sigma}} = \boldsymbol{\Sigma} - \sum_{m=1}^{M} w_m \boldsymbol{\Sigma}_m \tag{7}$$
$$= \sum_{m=1}^{M} w_m \int p(\boldsymbol{y}|m)(\boldsymbol{x_y}(m)-\boldsymbol{x_y})(\boldsymbol{x_y}(m)-\boldsymbol{x_y})^\top d\boldsymbol{y}$$

*where $\boldsymbol{\Sigma}$ is the global MMSE matrix with input distribution $p(\boldsymbol{x})$ and posterior mean $\boldsymbol{x_y}$, and $\boldsymbol{\Sigma}_m$ is the local MMSE matrix with input distribution $p(\boldsymbol{x}|m)$ and posterior mean $\boldsymbol{x_y}(m)$.*

*Proof.* Since $I(C;\boldsymbol{Y}) = h(\boldsymbol{Y}) - h(\boldsymbol{Y}|C) = I(\boldsymbol{X};\boldsymbol{Y}) - I(\boldsymbol{X};\boldsymbol{Y}|C)$ and $p(\boldsymbol{x}) = \sum_{m=1}^{M} w_m p(\boldsymbol{x}|m)$, according to Theorem 1, $\nabla_{\boldsymbol{\Phi}} I(C;\boldsymbol{Y})$ is equal to

$$\nabla_{\boldsymbol{\Phi}} I(\boldsymbol{X};\boldsymbol{Y}) - \nabla_{\boldsymbol{\Phi}} I(\boldsymbol{X};\boldsymbol{Y}|C) = \boldsymbol{R}\boldsymbol{\Phi}(\boldsymbol{\Sigma} - \sum_{m=1}^{M} w_m \boldsymbol{\Sigma}_m)$$

where $\boldsymbol{\Sigma}$ and $\boldsymbol{\Sigma}_m$ are the global and local MMSE matrix with input distribution $p(\boldsymbol{x})$ and $p(\boldsymbol{x}|m)$ respectively. From Bayes rule,

$$p(\boldsymbol{x}|\boldsymbol{y}) = \frac{p(\boldsymbol{x})p(\boldsymbol{y}|\boldsymbol{x})}{p(\boldsymbol{y})} = \frac{\sum_{m=1}^{M} w_m p(\boldsymbol{x}|m)p(\boldsymbol{y}|\boldsymbol{x})}{\sum_{m=1}^{M} w_m p(\boldsymbol{y}|m)}$$
$$= \frac{\sum_{m=1}^{M} w_m p(\boldsymbol{y}|m)p(\boldsymbol{x}|\boldsymbol{y},m)}{\sum_{m=1}^{M} w_m p(\boldsymbol{y}|m)} = \sum_{m=1}^{M} \widetilde{w}_m p(\boldsymbol{x}|\boldsymbol{y},m)$$

$$\widetilde{w}_m = p(m|\boldsymbol{y}) = \frac{w_m p(\boldsymbol{y}|m)}{\sum_{m'=1}^{M} w_{m'} p(\boldsymbol{y}|m')} = \frac{w_m p(\boldsymbol{y}|m)}{p(\boldsymbol{y})};$$
$$p(\boldsymbol{x}|\boldsymbol{y},m) = \frac{p(\boldsymbol{x}|m)p(\boldsymbol{y}|\boldsymbol{x})}{p(\boldsymbol{y}|m)};\ \boldsymbol{x_y}(m) = \int \boldsymbol{x}p(\boldsymbol{x}|\boldsymbol{y},m)d\boldsymbol{x}$$
$$\boldsymbol{x_y} = \int \boldsymbol{x}p(\boldsymbol{x}|\boldsymbol{y})d\boldsymbol{x} = \sum_{m=1}^{M} \widetilde{w}_m \boldsymbol{x_y}(m). \tag{8}$$

Consequently, we have

$$\boldsymbol{\Sigma} = \int p(\boldsymbol{y}) \int \sum_{m=1}^{M} \widetilde{w}_m p(\boldsymbol{x}|\boldsymbol{y},m)(\boldsymbol{x}-\boldsymbol{x_y})(\boldsymbol{x}-\boldsymbol{x_y})^\top d\boldsymbol{x}d\boldsymbol{y}$$
$$= \int \sum_{m=1}^{M} w_m p(\boldsymbol{y}|m)(\int p(\boldsymbol{x}|\boldsymbol{y},m)(\boldsymbol{x}-\boldsymbol{x_y}(m))(\boldsymbol{x}-\boldsymbol{x_y}(m))^\top d\boldsymbol{x}$$
$$+ (\boldsymbol{x_y}(m)-\boldsymbol{x_y})(\boldsymbol{x_y}(m)-\boldsymbol{x_y})^\top)d\boldsymbol{y}$$
$$= \sum_{m=1}^{M} w_m(\boldsymbol{\Sigma}_m + \int p(\boldsymbol{y}|m)(\boldsymbol{x_y}(m)-\boldsymbol{x_y})(\boldsymbol{x_y}(m)-\boldsymbol{x_y})^\top d\boldsymbol{y})$$

since $\boldsymbol{\Sigma}_m = \int p(\boldsymbol{y}|m) \int p(\boldsymbol{x}|\boldsymbol{y},m)(\boldsymbol{x}-\boldsymbol{x_y}(m))(\boldsymbol{x}-\boldsymbol{x_y}(m))^\top d\boldsymbol{x}d\boldsymbol{y}$ and $p(\boldsymbol{y})\widetilde{w}_m = w_m p(\boldsymbol{y}|m)$. Consequently, equation (7) is proved. □

The significance of Theorem 2 is that we obtain an explicit expression for the gradient of the MI objective function in (3) under *any* input signal distribution (1). Consequently, numerical optimization methods (*e.g.*, gradient descent) can be applied to solve (3). The equivalent MMSE matrix in (7) can be computed via Monte Carlo simulation and Bayesian inference (we discuss in Section 4 how we do this in practice). Our analytical result in Theorem 2 is very general, in the sense that we make no assumption on the signal distribution in each class. The algorithm can be summarized in the following steps:

1. Obtain the input signal distribution in (1) from training data. Initialize $\mathbf{\Phi}$.

2. Compute the equivalent MMSE matrix in (7) via Monte Carlo simulation.

3. Compute the gradient in (6) and update the projection matrix as $\mathbf{\Phi} \leftarrow \mathsf{orth}(\mathbf{\Phi} + \eta \nabla_{\mathbf{\Phi}} I(C; \boldsymbol{Y}))$, where $\eta$ is the step size and $\mathsf{orth}(\boldsymbol{A})$ means projecting $\boldsymbol{A}$ to an orthonormal matrix.

4. If converge, stop. Otherwise, go to step 2.

## 3. Theoretical Analysis

The orthonormal constraint on the projection matrix complicates the theoretical analysis of the optimal design. By relaxing this constraint and instead considering a power constraint we can leverage more results from communications and recent work in image reconstruction (Carson et al., 2012). The relaxed problem is

$$\max_{\mathbf{\Phi}} \quad I(C; \boldsymbol{Y}) \quad \text{s.t.} \quad \frac{1}{d} \operatorname{tr}\left(\mathbf{\Phi}\mathbf{\Phi}^\top\right) = 1 \tag{9}$$

where the trace constraint ensures that the rows of the projection matrix have on average unit-norm.

The following theorem characterizes the optimal projection matrix for the relaxed problem in terms of the singular value decompositions (SVD) of the noise covariance $\boldsymbol{R}^{-1} = \boldsymbol{U}_{\boldsymbol{R}}^\top \ \boldsymbol{D}_{\boldsymbol{R}}^{-1} \ \boldsymbol{U}_{\boldsymbol{R}}$ and the equivalent MMSE matrix $\widetilde{\mathbf{\Sigma}} = \boldsymbol{U}_{\widetilde{\mathbf{\Sigma}}} \ \boldsymbol{D}_{\widetilde{\mathbf{\Sigma}}} \ \boldsymbol{U}_{\widetilde{\mathbf{\Sigma}}}^\top$.

**Theorem 3.** *Given the measurement model in (2) and the multi-class signal model in (1), the projection matrix $\mathbf{\Phi}$ which optimizes the relaxed problem in (9) can be expressed via its SVD as $\mathbf{\Phi}^\star = \boldsymbol{U}_{\mathbf{\Phi}}^\star \ \boldsymbol{D}_{\mathbf{\Phi}}^\star \ \boldsymbol{V}_{\mathbf{\Phi}}^{\star\top}$ where $\boldsymbol{D}_{\mathbf{\Phi}}^\star$ is a square diagonal matrix of optimal singular values and the orthonormal matrices of optimal singular vectors are $\boldsymbol{U}_{\mathbf{\Phi}}^\star = \boldsymbol{U}_{\boldsymbol{R}}$ and $\boldsymbol{V}_{\mathbf{\Phi}}^\star = \boldsymbol{U}_{\widetilde{\mathbf{\Sigma}}^\star} \ \mathbf{\Pi}^\star$ for some optimal permutation matrix $\mathbf{\Pi}^\star$.*

*Proof.* From the KKT optimality conditions we know

$$\nabla_{\mathbf{\Phi}} \left\{ -I(C; \boldsymbol{Y}) - \eta \cdot \left[ 1 - \frac{1}{d} \operatorname{tr}\left(\mathbf{\Phi}\mathbf{\Phi}^\top\right) \right] \right\} \Bigg|_{\mathbf{\Phi}=\mathbf{\Phi}^\star} = -\boldsymbol{R} \ \mathbf{\Phi}^\star \widetilde{\mathbf{\Sigma}}^\star + 2\frac{\eta}{d} \cdot \mathbf{\Phi}^\star = 0$$

where the Lagrange multiplier $\eta \geq 0$, $\widetilde{\mathbf{\Sigma}}^\star$ is the equivalent MMSE matrix associated with the optimal projection matrix $\mathbf{\Phi}^\star$ and we have used the gradient result in Theorem 2. The optimal projection matrix must therefore also satisfy

$$2 \ \frac{\eta}{d} \cdot \mathbf{\Phi}^\star \mathbf{\Phi}^{\star\top} = \boldsymbol{R} \ \left( \mathbf{\Phi}^\star \widetilde{\mathbf{\Sigma}}^\star \mathbf{\Phi}^{\star\top} \right). \tag{10}$$

The left-hand side of this equation is symmetric and is diagonalized by $\mathbf{U}_{\mathbf{\Phi}}^{\star\top}$, which means that matrices $\boldsymbol{R}$ and $\mathbf{\Phi}^\star \mathbf{\Sigma}^\star \mathbf{\Phi}^{\star\top}$ commute and are simultaneously diagonalized by $\mathbf{U}_{\mathbf{\Phi}}^{\star\top}$. We can therefore write without loss of generality the optimal unitary matrices as

$$\boldsymbol{U}_{\mathbf{\Phi}}^\star = \boldsymbol{U}_{\boldsymbol{R}} \ \mathbf{\Pi}_{\boldsymbol{U}}^\star \boldsymbol{D}_{\boldsymbol{U}}; \quad \boldsymbol{V}_{\mathbf{\Phi}}^\star = \boldsymbol{U}_{\widetilde{\mathbf{\Sigma}}^\star} \mathbf{\Pi}_{\boldsymbol{V}}^\star \boldsymbol{D}_{\boldsymbol{V}}$$

where $\boldsymbol{D}_{\boldsymbol{U}}$ and $\boldsymbol{D}_{\boldsymbol{V}}$ are diagonal matrices with unit modulus diagonal elements, and $\mathbf{\Pi}_{\boldsymbol{U}}^\star$ and $\mathbf{\Pi}_{\boldsymbol{V}}^\star$ are permutation matrices. Noting that the action of the two permutation matrices can be captured by a single permutation matrix $\mathbf{\Pi}^\star$ and both mutual information and the MMSE matrix are independent of the unit modulus matrices, the result follows. □

The characterization in Theorem 3 of the projection matrix for the relaxed problem provides possible solutions to (3). For example, by setting the diagonal matrix $\boldsymbol{D}_{\mathbf{\Phi}}^\star$ to be the identity matrix we satisfy the orthonormal constraint on the projection matrix. This could be useful in the implementation of the gradient descent algorithm. For example,

- Theorem 3 takes the form of a fixed-point equation and could be used as stopping criteria in the proposed algorithm that indicates convergence.

- It is not known whether the mutual information is concave in $\mathbf{\Phi}$, Theorem 3 suggests an alternative (or extension) to gradient descent that could help avoid local optima.

- The projection matrix now consists of two rotation matrices, one of which always diagonalizes the noise which simplifies the calculation of the gradient.

A solution to the relaxed problem will necessarily be better than or equal to a solution to (3), since the constraint in (3) is a subset of that in (9) . Therefore the mutual information can be further improved by optimization over the singular values of the projection matrix. In the signal reconstruction scenarios in communications and image reconstruction, the mutual information is known to be concave in the squared singular values of the projection matrix when $\boldsymbol{U}_{\mathbf{\Phi}}^\star = \boldsymbol{U}_{\boldsymbol{R}}$ (Carson et al., 2012). This property can be used to give guarantees on convergence. However, the mutual information is not concave in this scenario. Nevertheless, by inserting the result for the optimal orthonormal matrices back into (10), the optimal squared singular values of the projection matrix satisfy

$$2 \ \frac{\eta}{d} \boldsymbol{D}_{\mathbf{\Phi}}^{2\star} = \boldsymbol{D}_{\mathbf{\Phi}}^{2\star} \ \boldsymbol{D}_{\boldsymbol{R}}^\star \ \left( \mathbf{\Pi}^{\star\top} \boldsymbol{D}_{\widetilde{\mathbf{\Sigma}}}^\star \mathbf{\Pi}^\star \right).$$

The equivalent MMSE matrix is a function of the projection matrix and therefore either $\left[\boldsymbol{D}_{\boldsymbol{\Phi}}^{2\star}\right]_{ii}$ is chosen to satisfy

$$2\,\frac{\eta}{d} = [\boldsymbol{D}_{\boldsymbol{R}}^{\star}]_{ii}\left[\boldsymbol{\Pi}^{\star\top}\boldsymbol{D}_{\tilde{\boldsymbol{\Sigma}}}^{\star}\boldsymbol{\Pi}^{\star}\right]_{ii}$$

or if no solution exists we choose $\left[\boldsymbol{D}_{\boldsymbol{\Phi}}^{2\star}\right]_{ii} = 0$. Note that from (7) in Theorem 2 we know that the equivalent MMSE matrix is positive semi-definite.

## 4. Mixture of GMMs Signal Model

In this section we focus on a specific signal input distribution, the mixture-of-GMMs signal model, in which signal from each class $m$ is modeled as a Gaussian Mixture Model (GMM), *i.e.*,

$$p(\boldsymbol{x}|m) = \sum_{o=1}^{O_m} \pi_{mo}\mathcal{N}(\boldsymbol{x};\boldsymbol{\mu}_{mo},\boldsymbol{\Omega}_{mo}) \qquad (11)$$

where $O_m$ is the number of Gaussian components for class $m$. As a result, the density in (1) reduces to

$$p(\boldsymbol{x}) = \sum_{m=1}^{M} w_m \sum_{o=1}^{O_m} \pi_{mo}\mathcal{N}(\boldsymbol{x};\boldsymbol{\mu}_{mo},\boldsymbol{\Omega}_{mo})$$

which is the mixture-of-GMMs signal model.

Under this specific signal model, the general Bayesian inference in (8) reduces to the inference of $\boldsymbol{x}$ under GMM priors. According to (Chen et al., 2010), the detailed Bayesian inference can be derived as

$$p(\boldsymbol{x}|\boldsymbol{y}) = \sum_{m=1}^{M} \widetilde{w}_m p(\boldsymbol{x}|\boldsymbol{y},m)$$

$$p(\boldsymbol{x}|\boldsymbol{y},m) = \sum_{o=1}^{O_m} \widetilde{\pi}_{mo}\mathcal{N}(\boldsymbol{x};\widetilde{\boldsymbol{\mu}}_{mo},\widetilde{\boldsymbol{\Omega}}_{mo})$$

$$\widetilde{\boldsymbol{\Omega}}_{mo} = (\boldsymbol{\Phi}^\top\boldsymbol{R}\boldsymbol{\Phi}+\boldsymbol{\Omega}_{mo}^{-1})^{-1};\ \widetilde{\boldsymbol{\mu}}_{mo} = \widetilde{\boldsymbol{\Omega}}_{mo}(\boldsymbol{\Phi}^\top\boldsymbol{R}\boldsymbol{y}+\boldsymbol{\Omega}_{mo}^{-1}\boldsymbol{\mu}_{mo})$$

$$p(\boldsymbol{y}|m) = \sum_{o'=1}^{O_m} \pi_{mo'}\mathcal{N}(\boldsymbol{y};\boldsymbol{\Phi}\boldsymbol{\mu}_{mo'},\boldsymbol{\Phi}\boldsymbol{\Omega}_{mo'}\boldsymbol{\Phi}^\top+\boldsymbol{R}^{-1})$$

$$\widetilde{\pi}_{mo} = \pi_{mo}\mathcal{N}(\boldsymbol{y};\boldsymbol{\Phi}\boldsymbol{\mu}_{mo},\boldsymbol{\Phi}\boldsymbol{\Omega}_{mo}\boldsymbol{\Phi}^\top+\boldsymbol{R}^{-1})/p(\boldsymbol{y}|m)$$

$$\widetilde{w}_m = \frac{w_m p(\boldsymbol{y}|m)}{\sum_{m'=1}^{M} w_{m'} p(\boldsymbol{y}|m')}.$$

The marginal density $p(\boldsymbol{y}) = \sum_{m'=1}^{M} w_{m'}p(\boldsymbol{y}|m')$ expressed in the denominator of $\widetilde{w}_m$ is also a mixture of GMM. The Matrix Inversion Lemma can be used to expedite the computations. Using the above equations, the equivalent MMSE matrix in (7) can be readily computed via Monte Carlo draws from $p(\boldsymbol{y})$, with $\boldsymbol{x}_{\boldsymbol{y}}$ and $\boldsymbol{x}_{\boldsymbol{y}}(m)$ provided by the above inference. Moreover, the inference naturally induces a Bayes classifier $\max_c p(c|\boldsymbol{y})$ where $p(c|\boldsymbol{y}) = \widetilde{w}_c$. We will use this mixture-of-GMMs signal model and the induced Bayesian classifier in the experiments.

## 5. Related Methods

Information-theoretic supervised dimensionality reduction was studied in (Torkkola, 2003; 2001). Instead of using Shannon entropy to define the mutual information, they used quadratic Rényi entropy to define a quadratic mutual information as

$I_T(C;\boldsymbol{Y}) = \sum_c \int (p(\boldsymbol{y},c) - p(\boldsymbol{y})p(c))^2 d\boldsymbol{y}$

where $p(c) = w_c$, and $p(\boldsymbol{y})$ is a mixture-of-GMMs defined in the same way as that in Section 4. The main advantage of using quadratic Rényi entropy is that the quadratic mutual information and its derivative can be expressed analytically for the GMM signal model without Monte Carlo simulations, due to the following property of Gaussian:

$$\int p(\boldsymbol{y}|m)p(\boldsymbol{y}|c)d\boldsymbol{y} = \int \sum_{o=1}^{O_m}\pi_{mo}\mathcal{N}(\boldsymbol{y};\boldsymbol{\Phi}\boldsymbol{\mu}_{mo},\boldsymbol{\Phi}\boldsymbol{\Omega}_{mo}\boldsymbol{\Phi}^\top+\boldsymbol{R}^{-1})$$

$$\times \sum_{r=1}^{O_c}\pi_{cr}\mathcal{N}(\boldsymbol{y};\boldsymbol{\Phi}\boldsymbol{\mu}_{cr},\boldsymbol{\Phi}\boldsymbol{\Omega}_{cr}\boldsymbol{\Phi}^\top+\boldsymbol{R}^{-1})d\boldsymbol{y}$$

$$= \sum_{o=1}^{O_m}\sum_{r=1}^{O_c}\pi_{mo}\pi_{cr}\mathcal{N}(\boldsymbol{0};\boldsymbol{\Phi}(\boldsymbol{\mu}_{mo}-\boldsymbol{\mu}_{cr}),\boldsymbol{\Phi}(\boldsymbol{\Omega}_{mo}+\boldsymbol{\Omega}_{cr})\boldsymbol{\Phi}^\top+2\boldsymbol{R}^{-1}).$$

In this paper, we will use a similar but different definition of quadratic mutual information

$$I_2(C;\boldsymbol{Y}) = h_2(\boldsymbol{Y}) - \sum_{m=1}^{M} w_m h_2(\boldsymbol{Y}|m)$$

where $h_2(\boldsymbol{Y}) = -\log\int p(\boldsymbol{y})^2 d\boldsymbol{y}$ is the quadratic Rényi entropy. This definition is more relevant to our Shannon entropy based approach, since by replacing $h_2(\boldsymbol{Y})$ with $h(\boldsymbol{Y})$, $I_2(C;\boldsymbol{Y})$ reduces to $I(C;\boldsymbol{Y})$. The optimization of $I_2(C;\boldsymbol{Y})$ is also straightforward, since the gradient can be expressed analytically due to the above property of Gaussians. We will compare this Rényi entropy based approach to our Shannon entropy based approach in the experiments. We emphasize that both $I_T(C;\boldsymbol{Y})$ and $I_2(C;\boldsymbol{Y})$ are approximations to $I(C;\boldsymbol{Y})$ for the sake of optimization, hence they cannot satisfy the bound in (4).

The Information Discriminant Analysis (IDA) (Nenadic, 2007) and Linear Discriminant Analysis (LDA) (Fisher, 1936) are derived under GMM signal model, which is a simplification and a special case of the mixture-of-GMMs signal model discussed in Section 4. It is interesting to compare our method with these two quantitatively. In the GMM signal model, the signal distribution in each class is modeled as a single Gaussian, i.e., $O_m = 1$ in (11) for all $m$. Hence $p(\boldsymbol{x}|m) = \mathcal{N}(\boldsymbol{x};\boldsymbol{\mu}_m,\boldsymbol{\Sigma}_m)$ and $p(\boldsymbol{x}) = \sum_{m=1}^{M} w_m\mathcal{N}(\boldsymbol{x};\boldsymbol{\mu}_m,\boldsymbol{\Sigma}_m)$. Thus the Bayesian inference in Section 4 can be further simplified. Under this simplified model assumption, the MI objective function in (3) can be expressed as

$$I(C;\boldsymbol{Y}) = h(\boldsymbol{Y}) - \sum_{m=1}^{M} w_m h(\boldsymbol{Y}|m)$$

$$= h(\boldsymbol{Y}) - \frac{1}{2}\sum_{m=1}^{M} w_m \log((2\pi e)^d \det(\boldsymbol{\Phi}\boldsymbol{\Omega}_m\boldsymbol{\Phi}^\top + \boldsymbol{R}^{-1})).$$

As illustrated above, $p(\boldsymbol{y})$ is also a GMM whose entropy cannot be expressed analytically. To overcome this problem, IDA approximates $h(\boldsymbol{Y})$ with a single Gaussian entropy with the same covariance matrix as the GMM $p(\boldsymbol{y})$, hence the objective function can be expressed as

$$I_{\text{IDA}}(C;\boldsymbol{Y}) = \frac{1}{2}\log((2\pi e)^d \det(\boldsymbol{\Phi}\boldsymbol{\Omega}\boldsymbol{\Phi}^\top + \boldsymbol{R}^{-1})) - \frac{1}{2}\sum_{m=1}^{M} w_m \log((2\pi e)^d \det(\boldsymbol{\Phi}\boldsymbol{\Omega}_m\boldsymbol{\Phi}^\top + \boldsymbol{R}^{-1}))$$

where $\boldsymbol{\Omega} = \sum_{m=1}^{M} w_m(\boldsymbol{\Omega}_m + (\boldsymbol{\mu}_m - \boldsymbol{\mu})(\boldsymbol{\mu}_m - \boldsymbol{\mu})^\top)$ is the prior covariance matrix for $\boldsymbol{x}$ and $\boldsymbol{\mu} = \sum_{m=1}^{M} w_m\boldsymbol{\mu}_m$ is the prior mean. Then the optimization of $I_{\text{IDA}}(C;\boldsymbol{Y})$ can be solved via gradient descent (Nenadic, 2007).

The LDA method (Fisher, 1936) simultaneously maximizes the between-class scattering and minimizes the within-class scattering of the projected data. It has been proven that under mild conditions this criterion is Bayes optimal (Hamsici & Martinez, 2008). The LDA criterion can be expressed as

$$I_{\text{LDA}}(C;\boldsymbol{Y}) = \frac{1}{2}\log((2\pi e)^d \det(\boldsymbol{\Phi}\boldsymbol{\Omega}\boldsymbol{\Phi}^\top + \boldsymbol{R}^{-1})) - \frac{1}{2}\log((2\pi e)^d \det(\boldsymbol{\Phi}(\sum_{m=1}^{M} w_m\boldsymbol{\Omega}_m)\boldsymbol{\Phi}^\top + \boldsymbol{R}^{-1})).$$

An analytical solution can be found for maximizing $I_{\text{LDA}}(C;\boldsymbol{Y})$, however the solution only permits the number of projections $d$ to be less than the class number $M$.

It is easy to prove that $I_{\text{IDA}}(C;\boldsymbol{Y}) \geq I(C;\boldsymbol{Y})$ (maximum entropy principle) (Nenadic, 2007) and $I_{\text{IDA}}(C;\boldsymbol{Y}) \geq I_{\text{LDA}}(C;\boldsymbol{Y})$ (concavity of $\log\det(\cdot)$). Clearly, only $I(C;\boldsymbol{Y})$ is the exact information-theoretic principle satisfying the Bayes error bound in (4), while the other two are approximations to the MI objective function. Another advantage of our method is that the higher-order information of the signal distribution is preserved in the objective function via $h(\boldsymbol{Y})$, while the other two methods only use first and second order statistics of the data. Even though $I(C;\boldsymbol{Y})$ cannot be expressed analytically, the optimization can still be done using the tool developed in Section 2.

# 6. Experiments

We test our method on three real datasets: Satellite, Letter and USPS. The first two are used in IDA (Nenadic, 2007) and can be downloaded from the UCI Machine Learning Repository. The third one is a standard digit recognition dataset with higher feature dimensions, which can also be downloaded from the Internet. A detailed description of the three datasets is as follows:

1. The 36-dimensional feature vectors in the Satellite data consist of pixel values of a $3\times3$ neighborhood in 4 spectral channels. The label for the central pixel belongs to one of the following six classes: real soil, cotton crop, grey soil, damp grey soil, soil with vegetation stubble and very damp grey soil. The training set contains 4435 samples, and the testing set contains 2000 samples.

2. The Letter data contains 16-dimensional feature vectors (statistical moments and edge counts) extracted from character images for the 26 capital letters (A to Z) with different fonts and random distortions. The training set has 16000 such stimuli and the testing set 4000.

3. The USPS data contains grey scale images of dimension $16 \times 16 = 256$ for handwritten digits $(0 \sim 9)$. There are 7291 training samples and 2007 testing samples.

The mixture-of-GMMs signal model is used, and the GMM density for each class is learned on the training data via the EM algorithm. Dirichlet Process (Blei & Jordan, 2006) GMM learning with variational Bayes inference was also tried to infer the mixture-of-GMMs model, yielding similar results. Two settings for the number of Gaussian components ($O_m$) are considered: $O_m = 1$ for all $m$, which reduces to the GMM signal model, and $O_m = 10$ for all $m$. The noise covariance matrix $\boldsymbol{R}^{-1}$ in (2) is set to be very small ($10^{-6}\boldsymbol{I}_d$). Four dimensionality reduction methods are considered: LDA, IDA, the quadratic Rényi entropy based method with objective function $I_2(C;\boldsymbol{Y})$, and the proposed Shannon entropy based method. For the proposed method, 2000 Monte Carlo particles are simulated to compute the equivalent MMSE matrix, and the step size for the gradient descent is set to be 0.01. The Bayes classifier $\max_c p(c|\boldsymbol{y})$ is employed using the learned signal model. The results are summarized in the following tables.

We observe that for all cases, the proposed method either gives the best performance, or is very near to the best. The state-of-art performance on the USPS data was obtained in (Tao et al., 2009). By adopting a nearest neighborhood rule-based classifier, they obtained classification accuracies of 0.7259, 0.8672, 0.8991 and 0.9182 using 3, 5, 7 and 9 designed projections respectively. Comparing to results in the table, we see that the proposed method is very competitive. Our strong result on this USPS dataset gives confidence to our method in general. The reason for our good performance is that we are directly maximizing $I(C; \boldsymbol{Y})$, which bounds the Bayes classification error $P_e$ through (4). The larger $I(C; \boldsymbol{Y})$ is, the smaller the upper bound of $P_e$ will be. All other objective functions ($I_{\text{LDA}}(C; \boldsymbol{Y})$, $I_{\text{IDA}}(C; \boldsymbol{Y})$ and $I_2(C; \boldsymbol{Y})$) are approximations to $I(C; \boldsymbol{Y})$, hence their performances are generally weaker.

*Table 1.* Classification accuracies on the Satellite data. The number in the parentheses is the number of Gaussian components ($O_m$) for each class.

| $d$ | LDA(1) | IDA(1) | RÉNYI(1) | PROPOSED(1) |
|---|---|---|---|---|
| 1 | 0.5650 | 0.6735 | 0.6880 | **0.7320** |
| 2 | 0.7835 | 0.7260 | 0.7860 | **0.8195** |
| 3 | 0.8415 | 0.8455 | 0.7955 | **0.8505** |
| 4 | **0.8470** | 0.8445 | 0.8170 | 0.8370 |
| 5 | **0.8445** | 0.8370 | 0.8200 | 0.8390 |
| $d$ | LDA(10) | IDA(10) | RÉNYI(10) | PROPOSED(10) |
| 1 | 0.5595 | 0.6725 | 0.7275 | **0.7390** |
| 2 | 0.7890 | 0.7380 | 0.8095 | **0.8325** |
| 3 | 0.8635 | **0.8725** | 0.8150 | 0.8675 |
| 4 | 0.8750 | 0.8695 | 0.8550 | **0.8805** |
| 5 | 0.8770 | 0.8780 | 0.8500 | **0.8880** |

*Table 2.* Classification accuracies on the Letter data.

| $d$ | LDA(1) | IDA(1) | RÉNYI(1) | PROPOSED(1) |
|---|---|---|---|---|
| 1 | 0.1812 | 0.1780 | 0.1812 | **0.1847** |
| 2 | **0.3785** | 0.3440 | 0.3485 | 0.3760 |
| 3 | 0.4715 | 0.4535 | **0.4965** | 0.4930 |
| 4 | 0.5715 | 0.5580 | 0.5580 | **0.6042** |
| 5 | 0.6285 | 0.6372 | 0.6425 | **0.6643** |
| 6 | 0.6927 | 0.6905 | 0.6900 | **0.7198** |
| 7 | 0.7210 | 0.7470 | 0.7073 | **0.7645** |
| 8 | 0.7515 | 0.7823 | 0.7470 | **0.7935** |
| $d$ | LDA(10) | IDA(10) | RÉNYI(10) | PROPOSED(10) |
| 1 | 0.1832 | 0.1895 | 0.2220 | **0.2273** |
| 2 | 0.4430 | 0.4183 | 0.4402 | **0.4698** |
| 3 | 0.5675 | 0.5330 | 0.6020 | **0.6490** |
| 4 | 0.6950 | 0.6723 | 0.6763 | **0.7675** |
| 5 | 0.7558 | 0.7575 | 0.7190 | **0.8315** |
| 6 | 0.8167 | 0.8170 | 0.6830 | **0.8740** |
| 7 | 0.8515 | 0.8760 | 0.7225 | **0.8988** |
| 8 | 0.8840 | **0.9150** | 0.8213 | 0.9123 |

We also observe that the performance using the mixture-of-GMMs signal model ($O_m = 10$) is generally better than that of the GMM signal model ($O_m = 1$), which is most obvious for the Letter dataset. This is because the mixture of GMM can model the data more precisely, which effectively improves the projection design and the Bayes classification.

*Table 3.* Classification accuracies on the USPS data.

| $d$ | LDA(1) | IDA(1) | RÉNYI(1) | PROPOSED(1) |
|---|---|---|---|---|
| 1 | 0.4694 | 0.3852 | 0.4654 | **0.5157** |
| 2 | 0.5994 | 0.4753 | 0.7354 | **0.7564** |
| 3 | 0.6761 | 0.5361 | 0.7947 | **0.8376** |
| 4 | 0.7967 | 0.5775 | 0.8371 | **0.8744** |
| 5 | 0.8555 | 0.6378 | 0.8605 | **0.8999** |
| 6 | 0.8819 | 0.7030 | 0.8809 | **0.9058** |
| 7 | 0.8894 | 0.7205 | 0.8814 | **0.9098** |
| 8 | 0.8889 | 0.7145 | 0.8789 | **0.9088** |
| 9 | 0.8939 | 0.7324 | 0.8899 | **0.9153** |
| $d$ | LDA(10) | IDA(10) | RÉNYI(10) | PROPOSED(10) |
| 1 | 0.4629 | 0.3852 | 0.4694 | **0.5227** |
| 2 | 0.6064 | 0.4983 | 0.7339 | **0.7623** |
| 3 | 0.6816 | 0.5725 | 0.7962 | **0.8505** |
| 4 | 0.8067 | 0.6403 | 0.8351 | **0.8804** |
| 5 | 0.8635 | 0.7000 | 0.8450 | **0.9033** |
| 6 | 0.8884 | 0.7534 | 0.8485 | **0.9188** |
| 7 | 0.8944 | 0.7683 | 0.8371 | **0.9183** |
| 8 | 0.9003 | 0.7723 | 0.7947 | **0.9198** |
| 9 | 0.9033 | 0.7828 | 0.7728 | **0.9287** |

The LDA and IDA method assume a GMM signal model ($O_m = 1$), hence the mixture of GMM signal model ($O_m = 10$) will not affect the projection design for LDA and IDA. However, as explained earlier, a finer signal model can help improve the classification performance. This is why we often observe a higher classification accuracy in LDA(10) (or IDA(10)) than that in LDA(1) (or IDA(1)), even though the designed projection matrices are exactly the same in the two cases; the brackets $(\cdot)$ indicate the number of GMM mixture components.

IDA performs poorly on the USPS data. This is because the global Gaussian approximation made in the $I_{\text{IDA}}(C; \boldsymbol{Y})$ objective function may not be appropriate for the USPS data with so much heterogeneity. The performance of the quadratic Rényi entropy based method is competitive, especially on the USPS dataset when $d \leq 4$. However, it is generally not as good as the proposed method, for reasons explained earlier. For all three datasets we also considered random projections designed based on draws from $\mathcal{N}(0, 1)$ with orthonormalization, and those were significantly worse than those of LDA, IDA, Renyi and the proposed method.

In summary, the performance of the proposed method is very promising. Its computational load is heavier than LDA and IDA, but the performance gain warrants the effort. Moreover, the projection design is done offline, so the testing speed will not be affected.

## 7. Conclusion

By harnessing a recent theoretical result on the gradient of MI with respect to the projection matrix (Palomar & Verdu, 2006), we have derived a new counterpart theorem for supervised dimensionality reduction. As a result, the Shannon MI objective function can be optimized directly without any approximation. We compared the proposed method to LDA, IDA and a quadratic Rényi entropy based method, both theoretically and empirically. Results on real datasets show the advantage of the proposed method. This study can be viewed as an example of how a research product from one area (communications theory) can benefit research in a seemingly different area (machine learning).

## Acknowledgement

The research reported here was supported by ARO, NGA, ONR and DARPA (KeCom program).